\pdfoutput=1

\documentclass{article}
\usepackage{spconf}
\usepackage{graphicx}
\usepackage{algorithm}
\usepackage{algpseudocode}
\usepackage{subfigure}
\usepackage{booktabs} 
\usepackage{hyperref}
\usepackage{amsmath}
\usepackage{multirow}
\usepackage[textsize=tiny]{todonotes}

\title{Partial Federated Learning}

\usepackage{times}
\usepackage{latexsym}

\name{Tiantian Feng$^1$, Anil Ramakrishna$^2$, Jimit Majmudar$^2$, Charith Peris$^2$, Jixuan Wang$^2$, Clement Chung$^2$,}
\secondname{\textit{Richard Zemel$^2$, Morteza Ziyadi$^2$, Rahul Gupta$^2$}}
\address{$^1$University of Southern California, $^2$Amazon Alexa AI}

\begin{document}

\maketitle

\begin{abstract}
    Federated Learning (FL) is a popular algorithm to train machine learning models on user data constrained to edge devices (for example, mobile phones) due to privacy concerns. Typically, FL is trained with the assumption that no part of the user data can be egressed from the edge. However, in many production settings, specific data-modalities/meta-data are limited to be on device while others are not. For example, in commercial SLU systems, it is typically desired to prevent transmission of biometric signals (such as audio recordings of the input prompt) to the cloud, but egress of locally (i.e. on the edge device) transcribed text to the cloud may be possible. In this work, we propose a new algorithm called Partial Federated Learning (PartialFL), where a machine learning model is trained using data where a subset of data modalities or their intermediate representations can be made available to the server. We further restrict our model training by preventing the egress of data labels to the cloud for better privacy, and instead use a contrastive learning based model objective. We evaluate our approach on two different multi-modal datasets and show promising results with our proposed approach.
\end{abstract}

\section{Introduction}
Existing FL paradigms typically assume a uniform set of restrictions applied to all available data modalities. With this framework of uniform restrictions, depending on available permissions on the data, we maybe allowed to move all of them to a server for \textit{central} model training or we maybe required to retain all data on the user devices (hereafter referred to as edge devices) for a \textit{federated} model training. However, in many real world settings, a subset of data modalities may carry looser constraints allowing easy egress of such data to a server.
As an example, consider a machine learning model built using medical data. In this case, certain biometric modalities such as the patient's speech recordings, electrocardiogram (ECG), electroencephalogram (EEG), continuous heart rate (HR) measurements are likely to be protected more strictly compared to anonymized doctor's notes and reports.
This is because biometric data are associated with Personal Identifiable Information (P.I.I) \cite{uludag2004biometric}, and sharing these can raise privacy concerns leading to a number of legislations to protect them, including the recently introduced California Biometric privacy law \cite{sbca} and Illinois Biometric Information Privacy Act \cite{bipa}.

Federated learning has been studied extensively in uni-modal settings while multi-modal federated learning has recently gained attention from several works \cite{chen2022fedmsplit, feng2023fedmultimodal}. However, existing learning paradigms do not support efficient and large scale training with distributed data modalities as described above, leading ML practitioners to choose the more restrictive setting of keeping all data on edge devices for federated model training. 
This brings with it all the challenges typical for federated learning: i) restricted model sizes due to small computational power on edge and ii) gradient drift issues due to heterogeneity in data, thus leading to lower performance compared to centralized training. 

In this work, we propose to address this gap by building a "Partial" Federated Learning model (PartialFL), where a model is trained using distributed data for which some modalities are shared centrally while other modalities and the class labels are only available on the edge device. In addition to the distributed training, a closely related objective is to utilize the shared modalities to improve on FL model performance by addressing the data heterogeneity challenge. Our approach is related to the existing paradigm of vertical federated learning \cite{yang2019federated}, where features for same set of samples maybe separated among different edge devices; in our case, we similarly have portions of data distributed between the central server and the edge devices, but the key difference in our work is that here we assume entirely different modalities of data exist on the edge and hence we can train a new artifact, such as an edge specific text encoder or acoustic signal encoder. Our main contributions in this work are:

\begin{itemize}
    \item We present a new Federated Learning algorithm to train a machine learning model when data modalities are split among different devices 
    \item We evaluate our algorithm on two different data sets present key results showing improvements in model performance
\end{itemize}
We also present additional results on a new dataset in the supplemental material along with various ablation results, highlighting the key areas of improvement introduced by our approach.
\vspace{-5mm}
\section{Approach}


In a multi-modal FL setting, we can categorize each data modality into either the protected \textbf{non-shareable group} or the less restricted \textbf{shareable group} based on whether they are permitted to be egressed to the central server. 
In this work, without loss of generalization, we assume that anonymized text data are shareable with the remote server (but note that our treatment holds for any other modality that is deemed shareable), while holding other modalities in the non-shareable group on the edge devices. For further protection against information leakage, we map the raw text data into latent representations using a pre-trained language models such as DistilBERT \cite{sanh2019distilbert}, and only share these representations with the server in all our experiments. 
The benefit of training with shareable data in the server is that we can now train a larger model due to increased availability of compute in the server, and also extract a more robust representation when compared to the model trained on an unbalanced local data set. We assume availability of labels on edge for local modal training, but in their absence we can leverage user feedback similar to \cite{sharma2022federated}, which we defer for future work.


\begin{figure*}[h]
    \centering
    \includegraphics[width=0.85\linewidth]{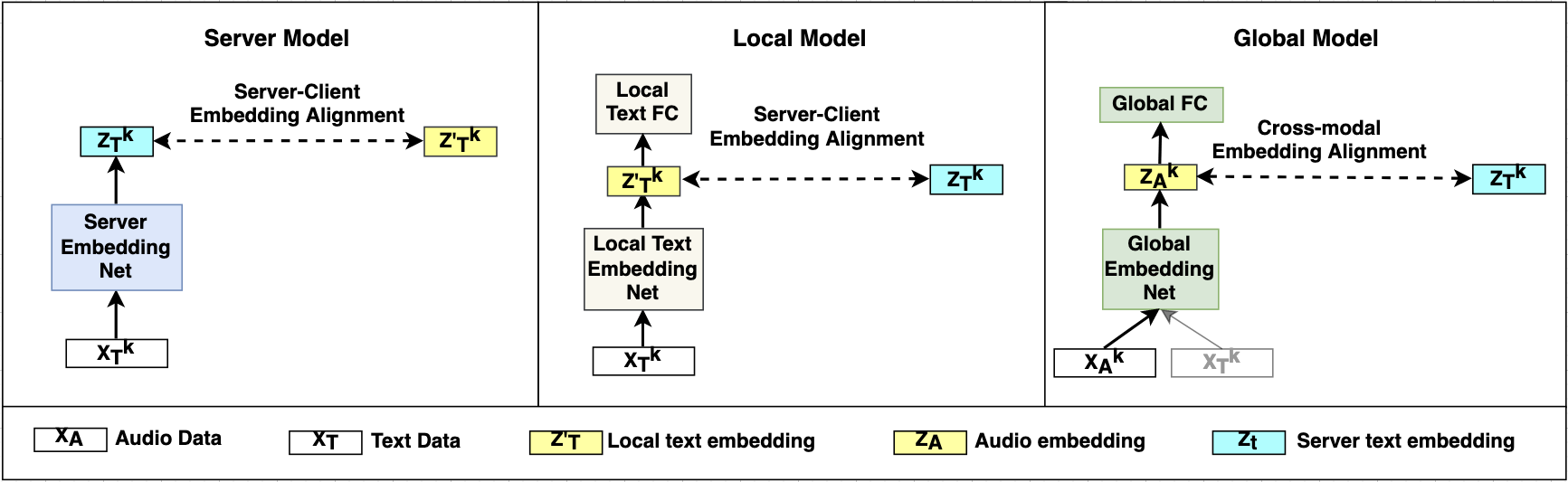}
    \caption{Different models in the PartialFL learning architecture.}
    \label{img:learning_architecture}
\end{figure*}

\vspace{-2mm}
\subsection{Outline}

We consider a decentralized setup with $K$ edge devices and a given multi-modal data set $\mathcal{D}$. For ease of exposition, we describe our approach using text and audio data, but note that it can be applied to any multi-modal setting. As noted in the previous section, latent representations of text data is assumed shareable but no form of audio data is shareable. We denote $\mathcal{D}^{k}: \{\mathbf{X}^{k}_{A,i}, \mathbf{X}^{k}_{T,i}, \mathbf{y}^{k}_{i}\}^{n_k}_{i=1}$ as the multi-modal dataset from $k^{th}$ edge device containing audio, text and labels respectively. $n_k$ is the size of the data set in $\mathcal{D}^{k}$, with $\mathcal{D} = \{\mathcal{D}^{1}, \mathcal{D}^{2}, ..., \mathcal{D}^{K}\}$ and the total number of utterances in $\mathcal{D}$ is $N=\sum^{K}_{k=1} n_k$. 

PartialFL follows a structure similar to regular FL but with one key modification: we maintain additional models on both the server and edge devices trained entirely on the shareable modality. We further augment model training by using cross device and cross-modality contrastive loss objectives. These are described in more detail below.

\vspace{-2mm}
\subsection{Learning Components}

Given the multi-modal setting described in the last section, the PartialFL framework consists of three model components as shown in Figure \ref{img:learning_architecture}: the server model $\mathcal{F}_{s}(\cdot)$, the global model $\mathcal{F}_{g}(\cdot)$, and the local models $\mathcal{F}_{k}(\cdot)$ where $k\in\{1, 2, ..., K\}$.

\noindent \textbf{The server model} $\mathcal{F}_{s}(\cdot)$ exists in the server as an encoder which is trained on the shareable data ($\mathbf{X_{T}}$) to generate embeddings. In our example, we define the server-side textual embedding as $\mathbf{z_{T}}$. Since we do not have training labels at the server to train a classifier, the learning objective of $\mathcal{F}_{s}(\cdot)$ is to reduce the distance between the server generated textual embedding $\mathbf{z_{T}}$ with the local textual embedding $\mathbf{z'_{T}}$.

\noindent \textbf{The global model} $\mathcal{F}_{g}(\cdot)$ is a typical FL model and is trained in a distributed manner on the non-shareable data on edge, followed by a global aggregation in the server using typical FL algorithms, like FedAvg \cite{konevcny2016federated} or FedProx \cite{li2020federated}. 
The global model can be either audio only or a multi-modal global model. 
The objective is to learn $\mathcal{F}_{g}(\cdot)$ parameterized by $\mathcal{\theta}_{g}$ over the data set $\mathcal{D}$ without accessing $\mathbf{X_{A}}$ from edge devices. Since not every edge device may have every modality, training the global model can further help those devices with missing modalities.

Further, since the edge-side training of this model can suffer from gradient drifts \cite{karimireddy2020scaffold}, we add a cross-modal alignment objective to decrease the distance between the audio embedding $\mathbf{z_{A}}$ (or multi-modal embedding $\mathbf{z_{M}}$) and $\mathbf{z_{T}}$. This idea is similar to the model contrastive loss presented in \cite{li2021model}. 

\noindent \textbf{The local model} $\mathcal{F}_{k}(\cdot)$ is only available in the $k^{th}$ edge device. Unlike the server model $\mathcal{F}_{s}(\cdot)$, here we have access to data labels so the local model includes a classifier trained using cross-entropy loss over the text modality. Note that the full $\mathcal{F}_{k}(\cdot)$ is not shareable since the server would then be able to infer local labels using $\mathcal{F}_{k}(\cdot)$ as $\mathbf{X_{T}}$ is already uploaded to the server. Instead, the edge device only sends the local textual embeddings $\mathbf{z'_{T}}$ generated from the encoder layer of $\mathcal{F}_{k}(\cdot)$. Similar to training the global model, over-fitting can occur easily while training this model, so we add an embedding alignment loss to minimize the distance between the local embeddings $\mathbf{z'_{T}}$ from the server generated embeddings $\mathbf{z_{T}}$.

The key intuition behind PartialFL is that by iteratively aligning server side representations of the shared text modality ($\mathbf{z_{T}}$) with different client side representations $\mathbf{z'_{T}}$ on the server, and by aligning the edge side text and audio representations with the realigned $\mathbf{z_{T}}$, we are improving model robustness to extreme data heterogeneity.

\vspace{-2mm}
\subsection{Cross-modal Alignment}

Cross-modal alignment is a popular learning task when working with multiple modalities \cite{gabeur2020multi, ging2020coot}. This learning task focuses on aligning embeddings of the same instance across the different modalities. More concretely, we compute modality specific representations denoted as $\mathbf{z_{A}}$, $\mathbf{z_{T}}$ and $\mathbf{z_{M}}$ for the audio embedding, text embedding and the multi-modal embedding respectively. Cross-modal alignment in PartialFL aims to push these embeddings close to each other if they belong to the same data sample, and increase the distance between them otherwise.

Similar to previous work \cite{zolfaghari2021crossclr}, a \textbf{positive pair} is defined as $\mathbf{z_{A}}$ (or $\mathbf{z_{M}}$) and $\mathbf{z_{T}}$ from the same data sample. On the other hand, \textbf{negative pairs} are constructed from different data samples. More precisely, we define \textbf{intra-modal negative pairs} as the embeddings from the same modality but different data samples. For instance, $(\mathbf{z}^{k}_{A,i}, \mathbf{z}^{k}_{A,j})$ is a negative pair if $i \neq j$. Further, we define \textbf{inter-modality negative pairs} as embeddings from different modalities \textit{and} data samples. An example of the inter-modality negative pair is $(\mathbf{z}^{k}_{A,i}, \mathbf{z}^{k}_{T,j})$ where $i \neq j$. With the audio global model, we can define the loss term $\mathcal{L}^{T \rightarrow A}_{i}$ for data $\mathbf{X}^{k}_{i}$ in a training batch of size B with a temperature parameter $\tau$ as:

\footnotesize
\begin{equation}
    \mathcal{L}^{T \rightarrow A}_{i} = -\log \frac{e^{(\mathbf{z}^{k}_{A,i})^{T}\mathbf{z}^{k}_{T,i}/\tau}}{\sum^{B}_{j=1,i\neq j} e^{(\mathbf{z}^{k}_{A,i})^{T}\mathbf{z}^{k}_{A,j}/\tau} + e^{(\mathbf{z}^{k}_{A,i})^{T}\mathbf{z}^{k}_{T,i}/\tau}}
\end{equation}
\normalsize

\subsection{Embedding Alignment at Server and Local models}

As noted before, to prevent over-fitting, we aim to decrease the distance between the local and server side textual embeddings $\mathbf{z'_{T}}$ and $\mathbf{z_{T}}$ using the contrastive loss. We define a positive pair as $\mathbf{z'_{T}}$ and $\mathbf{z_{T}}$ from the same data sample, and a negative pair from different samples. We can then write the server side contrastive loss $\mathcal{L}^{L \rightarrow S}_{i}$ as: 

\footnotesize
\begin{equation}
    \mathcal{L}^{L \rightarrow S}_{i} = -\log \frac{e^{(\mathbf{z}^{k}_{T,i})^{T}\mathbf{z'}^{k}_{T,i}/\tau}}{\sum^{B}_{j=1,i\neq j} e^{(\mathbf{z}^{k}_{T,i})^{T}\mathbf{z}^{k}_{T,j}/\tau} + e^{(\mathbf{z}^{k}_{T,i})^{T}\mathbf{z'}^{k}_{T,i}/\tau}}
\end{equation}
\normalsize

We define the edge side loss $\mathcal{L}^{S \rightarrow L}_{i}$ similarly. 


\subsection{Learning Algorithm}

Unlike training centralized contrastive models, PartialFL needs to optimize the server model and the edge models asynchronously. To optimize the edge model and the server model in such manner, we use alternating minimization (AM) similar to FedGKT \cite{he2020group}, where we alternatively fix one model and optimize the other. The full algorithm is presented in Appendix \ref{appendix: learning algorithm}.

\vspace{-2mm}
\subsection{Implementation}

We use PyTorch to implement the PartialFL and the other baselines. In this study, we experiment with two popular FL algorithms: FedAvg and FedProx. We fix the weight of the proximal loss in FedProx as 0.01. 
We fix the number of edge devices as 200. When experimenting with the Food-101 data set, we choose the edge sample rate in each training round as 10\%. On the other hand, we regard each speaker as a separate client in the emotion recognition task as it provides a natural data split in the FL. Since there are fewer clients in emotion recognition data sets, we decide to use an edge sample rate of 50\% in each global training round. 

\vspace{-2mm}
\subsection{Hyper-parameters}
 We set the local training batch size as 16 and the local training epoch as 1 in all FL algorithms. We set the learning rate as 0.0001 and 0.0005 in training the emotion recognition task and Food-101 classification tasks, respectively. We apply the Adam optimizer in all experiments. The global training round is 150 in the emotion recognition task and 200 in the Food-101 data set. In PartialFL, we explore the temperature parameter $\tau \in \{0.05, 0.1, 0.2\}$. We tune the weight $\beta$ in both $\mathcal{L}_{glob}$ and $\mathcal{L}_{loc}$ in $\{0.001, 0.01\}$ in training emotion recognition models.

\vspace{-4mm}
\section{Experiments}
We evaluate the PartialFL algorithm on two datasets from the Speech Emotion Recognition task (SER): IEMOCAP \cite{busso2008iemocap} which contains utterances from ten subjects expressing various categorical emotions, and MSP-Improv \cite{busso2016msp} which contains multi-modal emotion recognition data set collected from 12 speakers. Additional details on all our datasets and models are provided in Appendix \ref{sec:experiment_details}. We also present results on image classification task from the Food101 dataset along with detailed ablation studies in Appendix \ref{sec:additionalresults}. 

\begin{figure*}[h]
    \centering
    \includegraphics[width=0.85\linewidth]{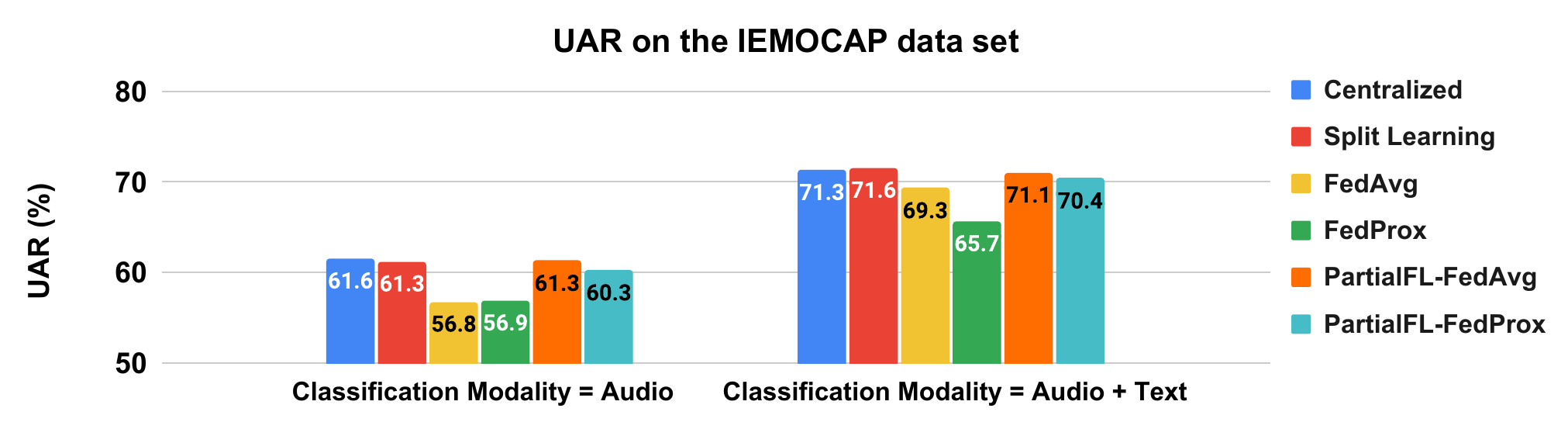}
    \includegraphics[width=0.85\linewidth]{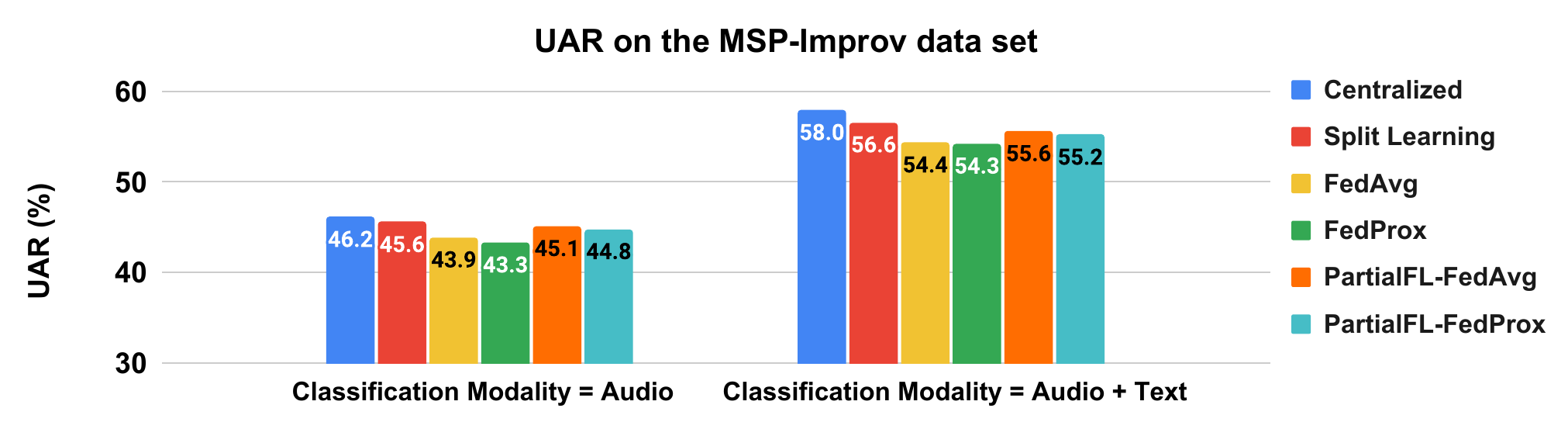}
    \caption{Model performance for the IEMOCAP and MSP-Improv data set. PartialFL considerably outperforms FL and SL and has performance close to the Centralized upper bound.}
    \label{img:ser_baseline}
\end{figure*}

\subsection{Baselines}

\subsubsection{Centralized} Here we assume that all data modalities are available in one central server for training in both uni-modal and multi-modal settings. Since we are no longer limited by edge size computational power, this acts as an upper limit for the model performance. 

\vspace{-3mm}
\subsubsection{Split Learning} We also implemented split learning to compare against our model. We set model size to be same as in the centralized baseline, and it serves as an upper bound for decentralized training. However, split learning incurs significant communication costs in practice compared to typical federated learning (refer to Section 3.3 in \cite{thapa2022splitfed} for a comparison of training times for both algorithms). 

\vspace{-2mm}
\subsubsection{Federated Learning} Finally, we also compare against a typical FL baseline. We experiment with both FedAvg and FedProx \cite{li2020federated} variants of FL, and set the global model size to be same as PartialFL for a fair comparison. 

\vspace{-3mm}
\section{Results}

We report unweighted average recall (UAR) score to measure the model performance. We use each recording session as a separate test fold, and repeat the training 5 and 6 times on the IEMOCAP and the MSP-Improv data sets respectively. 

\vspace{-2mm}
\subsection{Uni-model global model (audio only)}
Full results from our experiments are shown in Figure \ref{img:ser_baseline}; PartialFL showed stronger results than both FedProx and FedAvg in both datasets, but especially so in IEMOCAP, where we observed nearly 4.00\% improvement. 
Further, PartialFL approaches the performance of the centralized and split learning baselines by leveraging the additional data modality leading to improved robustness, while still retaining the benefits of federated learning.

\vspace{-2mm}
\subsection{Multi-modal global model (audio+text)}

In this setting, we observed stronger performance in the centralized and SL baselines, with SL surprisingly outperforming centralized training in both data sets. However, similar to the previous experiments, PartialFL consistently outperforms both FL baselines, with overall improvement in the range of 1.0-2.0\%, highlighting the robustness of our proposed approach. 

\vspace{-5mm}
\section{Conclusion}

We propose a novel Federated Learning framework called PartialFL with a goal of addressing the heterogeneity challenge in FL and improve the final model performance. Unlike traditional FL, PartialFL allows some modalities to be shared with the server which enables us to train a robust embedding network over the shared modality in the server. 
We experiment with multiple multi-modal data sets and report strong performance against three baselines. We observe that PartialFL consistently outperforms traditional FL in all tasks, and approaches the performance of centralized models, as well as split learning without any of its communication overhead or straggler problems. 
Future work includes deploying PartialFL in real world applications to further evaluate its efficacy.  

\bibliography{references}
\bibliographystyle{IEEEbib}

\newpage
\appendix

\section{Limitations and Future Work}
\label{sec:limitations}
The primarily limitation of our proposed approach is the added computation needed to train local and server models when compared to typical FL model training. Further, since we use pretrained models to extract sentence and image representations, we maybe inadvertently exposed to any underlying dataset biases in these models. We encourage downstream applications of this approach to suitably evaluate the model performance against different demographic cohorts before deployment. 

Another area of concern is the privacy risk associated with sharing the representation of a modality to the remote server. Many recent studies have demonstrated that FL is vulnerable to privacy attacks such as data reconstruction attacks \cite{zhu2019deep}, label inference attack \cite{fu2022label}, and property inference attack \cite{feng2021attribute}. In our experiments, we employ preliminary strategies to protect privacy by sharing representations of the shareable modality instead of raw data as well as only sharing representations from modalities without any biometric information (text). Further, we also place a strong constraint in our model training whereby we do not share class labels with the server. Nevertheless, it is still possible that some of aforementioned privacy attacks can achieve high success rates using just the shareable representations. For instance, the privacy attacker may formulate contrastive learning objectives between the model updates and shared representations to implement more effective inference attacks. Our future works plan to extensively investigate the privacy risks associated with sharing modalities in FL.

\section{Related Work}
Several algorithms have been developed to train with distributed data with disjoint modalities stored on different devices, but each of these have specific drawbacks which we highlight here. 

While we can assume that all modalities are restricted to the edge devices and use traditional FL, this poses challenges for training on low power devices and in cases of high data heterogeneity. \textit{Split learning} (SL) \cite{gupta2018distributed} is one way to address these concerns by splitting the entire ML model into multiple smaller parts and distributing them on both the central server and edge devices. Each model part can further be constrained to a specific data modality. In a typical SL setting, the first edge device initializes the training by performing forward propagation on local data, and then uploads the \textit{smashed} data (outermost activations of the local model) and ground-truth labels to the central server. In the next step, the central server continues forward propagation from the \textit{smashed} data using a server-side model. Then, the server starts back-propagation on its model and sends the gradients at the cut layer back to the edge for local back-propagation. After finishing back-propagation at the first edge device, it shares the trained model to the next edge device, and the training process continues. While this theoretically allows us to train large multi-modal models, the main drawback here is the sequential training process which makes it considerably slower than FL. Further, there is a significant communication overhead for each model update, since we need to propagate the intermediate activations from edge to server in the forward pass and then back in the backward pass, \textit{for each data sample and on all the edge devices}. In addition, SL can also suffer from server straggler problem as the training process requires frequent communications between the server and the edge devices. In our work, we avoid all of these problems by training device specific encoders until convergence which are then shared with other devices/the centralized server more efficiently (in terms of both communication cost and time). We assume availability of labels on edge for local modal training, but in their absence we can leverage user feedback similar to \cite{sharma2022federated}.

Federated Group Knowledge Transfer (FedGKT) \cite{he2020group} was originally proposed to address the challenges of training with non-IID datasets; it can also be applied to train a large multi-modal model in a distributed manner since it allows us to train device/modality specific encoders till convergence which are then aggregated centrally using knowledge distillation. However, FedGKT assumes that edge specific labels are also shared with the server which enables them to train a server side classifier till convergence, but these labels may not always be available, or not permitted to be shared freely with the server. For example, patients may prefer not to share their diagnosis information (class labels) outside the hospital. In our work, we therefore avoid label sharing entirely by using a contrastive objective to train the device specific encoders, and show strong model performance without sharing any edge specific labels with the central server. 

Split Federated Learning (SFL) \cite{thapa2022splitfed} is a recent work which attempts to bridge SL and FL. However, this also carries heavy communication costs due to the combined overheads of split learning and federated learning, which limits its applications to practical settings. Further, in our experiments, they do not demonstrate any gains in accuracy compared to our baselines.  

Finally, while federated learning has been studied extensively in uni-modal settings, multi-modal federated learning has recently gained attention from several works \cite{chen2022fedmsplit, xiong2022unified, yu2023multimodal, feng2023fedmultimodal}. 

\section{Learning Algorithm}
\label{appendix: learning algorithm}
Detailed training steps are as follows:

\noindent \textbf{Step 1 (server):} At the beginning of each global round, the server samples the edge devices and sends the global model $\mathcal{F}_{g}(\cdot)$ parameterized by $\mathcal{\theta}_{g}$ and $\mathbf{z_T}$ to each device.

\noindent \textbf{Step 2 (edge):} After receiving $\mathcal{\theta}_{g}$ and the server side embeddings $\mathbf{z}^{k}_{T}$, the edge device trains the global $\mathcal{F}_{g}(\cdot)$ on its local image data $\mathbf{X_{I}}$. The learning objective is a combination of cross-entropy loss $\mathcal{L}_{CE}$ and the cross-modal contrastive loss $\mathcal{L}^{T \rightarrow I}$. We use the weight parameter $\beta$ to set the importance of the cross-modal contrastive loss. The combined loss for the image global model is shown below:
\begin{equation}
    \mathcal{L}_{glob} = \mathcal{L}_{CE}(\mathcal{F}_{g}(\theta_{g}; \mathbf{X}^{k}_{I}), y^{k}) + \beta \mathcal{L}^{T \rightarrow I}(\mathbf{z_{I}}, \mathbf{z_{T}})
\end{equation}
\noindent \textbf{Step 3 (edge):} Next, the edge device trains its local model using $\mathbf{X_{T}}$. 
Here, the training objective is a weighted sum between the cross-entropy loss $\mathcal{L}_{CE}$ and the contrastive loss $\mathcal{L}^{S \rightarrow L}$. 
\begin{equation}
    \mathcal{L}_{loc} = \mathcal{L}_{CE}(\mathcal{F}_{k}(\theta_{k}; \mathbf{X}^{k}_{T}), y^{k}) + \beta \mathcal{L}^{S \rightarrow L}(\mathbf{z'_{T}}, \mathbf{z_{T}})
\end{equation}
\noindent \textbf{Step 4 (server):} Finally, the server receives the trained global model and edge generated embeddings $\mathbf{z'_{T}}$. The server then aggregates the global model using weighted average and also trains the server side model $\mathcal{F}_{s}(\cdot)$ using the contrastive loss $\mathcal{L}^{L \rightarrow S}$:


We summarize the full training algorithm in Algorithm \ref{alg:partialfl} and a pictorial representation of the same in Figure \ref{img:cross_modal_framework} in Appendix \ref{sec:appendix_a}.

Figure \ref{img:cross_modal_framework} shows a pictorial representation of our full learning algorithm, which is listed in detail in Algorithm \ref{alg:partialfl}.

\label{sec:appendix_a}
\begin{figure*}[h]
    \centering
    \includegraphics[width=0.95\linewidth]{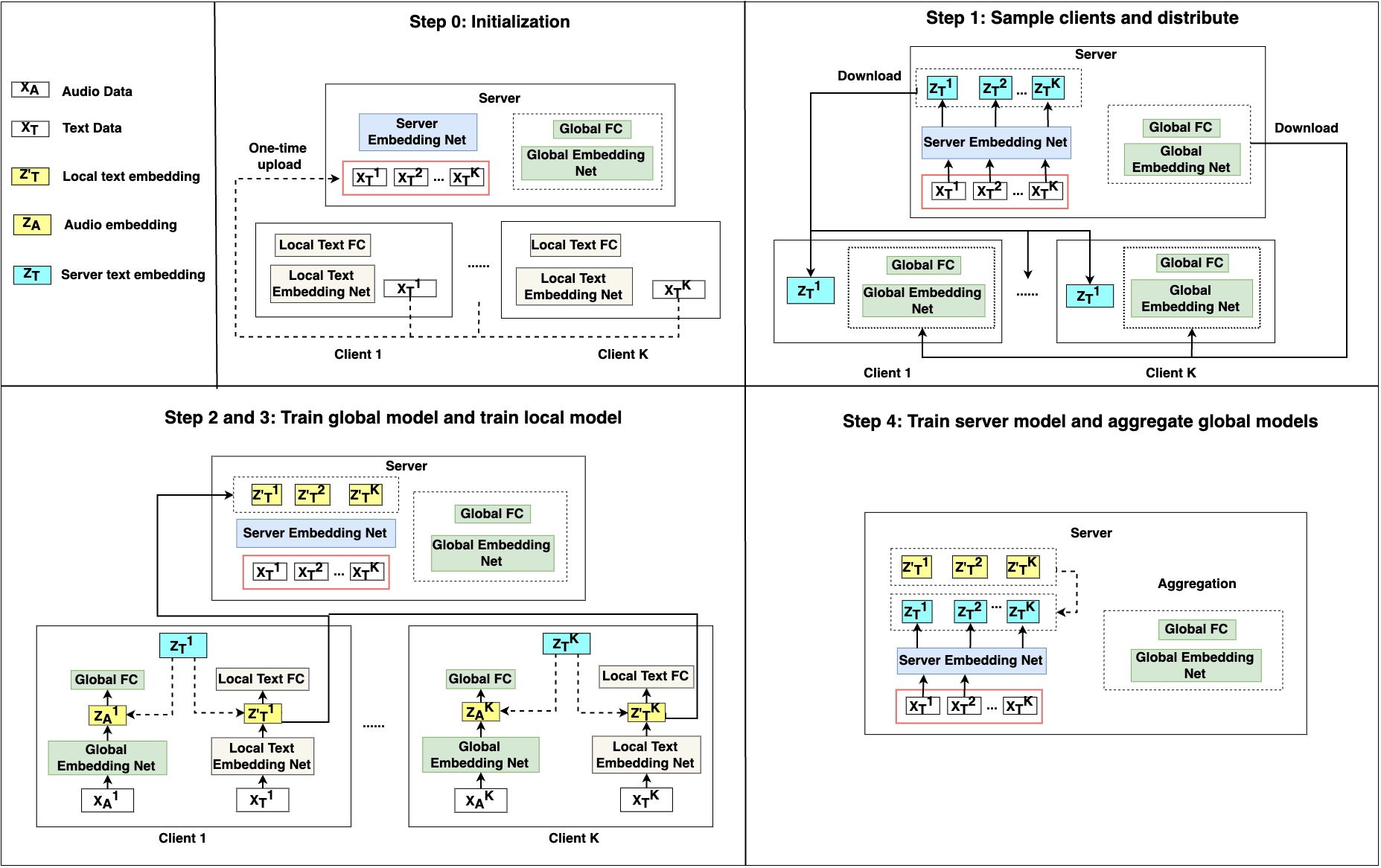}
    \caption{Training steps of the PartialFL framework. Training steps 1, 2, 3, and 4 are repeated in each global training round.}
    \label{img:cross_modal_framework}
\end{figure*}

\begin{algorithm}
    \caption{\textbf{PartialFL}}
    \begin{algorithmic}[1]
        
        \State{\textbf{Server Initialize: }}{$\mathbf{\theta}^{0}_{g}$, $\mathbf{\theta}_{s}$}
        \For{$k \in \{1, 2, ..., K\}$}
            \State{\textbf{Client Initialize: }}{$\mathbf{\theta}_{k}$ }
            \State{\textbf{Upload $\mathbf{X^{k}_{T}}$ to server}}
        \EndFor
        
        \State{\textbf{Server Executes:}}
        \For{$k \in \{1, 2, ..., K\}$}
            \State $\mathbf{z^{k}_{T}} \gets \mathcal{F}_{s}(\mathbf{\theta}_{s}; \mathbf{X^{k}_{T}})$
        \EndFor
        \For{Each round $t=0,...,T-1$}
            \State \textbf{\#\# Step 1: Sample and distribution}
            \State{Sample edge devices $\mathcal{S}\in\{1, 2, ..., K\}$}
            \State{Distribute $\mathbf{\theta}^{t}_{g}$ and $\mathbf{z^{k}_{T}}$}
            
            \State \textbf{\#\# Step 2 and 3: Client training}
            \For{Each edge device $k \in \mathcal{S}$ in parallel}
                \State $\mathbf{\theta}^{t}_{g,k}, \mathbf{\theta}_{k} \gets $  \textbf{ClientTrain($\mathbf{\theta}^{t}_{g}$,$\mathbf{\theta}_{k}$,$\mathbf{z^{k}_{T}}$,$\mathcal{D}^{k}$)}
                
                \State $\mathbf{z'^{k}_{T}} \gets \mathcal{F}_{k}(\mathbf{\theta}_{k}; \mathcal{D}^{k})$
                \State{\textbf{Upload $\mathbf{\theta}^{t}_{g, k}, \mathbf{z'^{k}_{T}}$ to server}}
            \EndFor
            \State \textbf{\#\# Step 4.1: Server trains $\mathbf{\theta}_{s}$}
            \For{Each edge device $k \in \mathcal{S}$}
                \State $\mathbf{\theta}_{s} \gets $  \textbf{ServerTrain($\mathbf{\theta}^{t}_{s}$, $\mathbf{z'^{k}_{T}}$, $\mathbf{X}^{k}_{T}$)}
            \EndFor
                
            \For{$k \in \{1, 2, ..., K\}$}
                \State $\mathbf{z^{k}_{T}} \gets \mathcal{F}_{s}(\mathbf{\theta}_{s}; \mathbf{X^{k}_{T}})$
            \EndFor
            
            \State \textbf{\#\# Step 4.2: Server aggregates $\mathbf{\theta^{t}_{g,k}}$}
            \State $\mathbf{\theta^{t+1}_{g}} \gets \frac{1}{|\mathcal{S}|} \sum_{k\in \mathcal{S}} \mathbf{\theta_{g,k}^{t}}$
            
        \EndFor

        \Function{\textbf{ClientTrain($\mathbf{\theta}_{g}$, $\mathbf{\theta}_{k}$, $\mathbf{z_{T}}$, $\mathcal{D}$)}}{}
            
            \For{Local epoch $e$ from 0 to $E-1$}
                \For{Iteration $i$ from 0 to $I-1$}
                    \State Sample mini-batch $\mathbf{l}$ from $\mathcal{D}$
                    
                    \State $\mathbf{\theta}_{g} \gets \mathbf{\theta}_{g}-\eta\nabla_{\theta_{g}}\mathcal{L}_{glob}(\mathcal{\theta}_{g}; \mathcal{D^{\mathbf{l}}, \mathbf{z^{l}_{T}}})$
                    
                    \State $\mathbf{\theta}_{k} \gets \mathbf{\theta}_{k}-\eta\nabla_{\theta_{k}}\mathcal{L}_{local}(\mathcal{\theta}_{k}; \mathcal{D^{\mathbf{l}}}, \mathbf{z}^{\mathbf{l}}_{T})$
                \EndFor
            \EndFor
            
        \State \textbf{return} {$\mathbf{\theta}_{g}$, $\mathbf{\theta}_{k}$, $\mathbf{z'}_{T}$}
        \EndFunction
        
        \Function{\textbf{ServerTrain($\mathbf{\theta}_{s}$, $\mathbf{z'_{T}}$, $\mathbf{X_{T}}$)}}{}
            
            \For{Iteration $i$ from 0 to $I-1$}
                \State Sample mini-batch $\mathbf{l}$ from $\mathbf{X_{T}}$
                
                \State $\mathbf{\theta}_{s} \gets \mathbf{\theta}_{s}-\eta\nabla_{\theta_{s}}\mathcal{L}_{server}(\mathcal{\theta}_{s}; \mathbf{X^{l}_{T}}, \mathbf{z'^{l}_{T}})$
                
            \EndFor
        \State \textbf{return} {$\mathbf{\theta}_{s}$}
        \EndFunction
    \end{algorithmic}
    \label{alg:partialfl}
\end{algorithm}

\section{Experiment Details}
\label{sec:experiment_details}

\subsection{Dataset}
\subsubsection{IEMOCAP}
The IEMOCAP database \cite{busso2008iemocap} was collected using multi-modal sensors to capture motion, audio, and video of human interactions. The original corpus contains 10,039 utterances from ten subjects (five male and five female) expressing various categorical emotions from improvised and scripted scenarios. 
Following \cite{zhang2018attention}, we focus on the improvised sessions. We use the four most frequent emotion labels: neutral, sad, happiness, and anger for training the SER model due to the data imbalance in other labels.

\subsubsection{MSP-Improv}
The MSP-Improv corpus \cite{busso2016msp} is a multi-modal emotion recognition data set captured from improvised scenarios. The data is collected from 12 speakers (six male and six female) and includes audio and textual data of utterances spoken in different recording conditions. 

\subsubsection{UPMC Food101}

The UPMC Food101 dataset \cite{wang2015recipe} consists of web pages with textual recipe descriptions for 101 food labels automatically retrieved online. Each page was matched with a single image, where the images were obtained by querying Google Image Search for the given category. Examples of food labels are Filet Mignon, Pad Thai, Breakfast Burrito and Spaghetti Bolognese. The web pages were processed with html2text\footnote{github.com/aaronsw/html2text} to obtain the raw text.

\subsection{Data Statistics}
Data statistics are shown in Table \ref{tab:dataset}.
\begin{table}
    
    \centering
    \caption{Statistics of data sets used in this study.}
    
    \begin{tabular}{p{1.25cm}p{1.25cm}p{1.25cm}p{1.25cm}p{1.25cm}}
        
        \toprule
        \multicolumn{1}{c}{} & 
        \multicolumn{1}{c}{\textbf{Train}} & 
        \multicolumn{1}{c}{\textbf{Dev}} &
        \multicolumn{1}{c}{\textbf{Test}} &
        \multicolumn{1}{c}{\textbf{}}
        \rule{0pt}{2ex} \\ \midrule
        
        \multicolumn{1}{l}{\textbf{Food101}} & 
        \multicolumn{1}{c}{58545} &
        \multicolumn{1}{c}{6556} &
        \multicolumn{1}{c}{21695} &
        \multicolumn{1}{c}{} \rule{0pt}{2ex} \\

        \toprule
        \multicolumn{1}{c}{} & 
        \multicolumn{1}{c}{\textbf{Neu.}} & 
        \multicolumn{1}{c}{\textbf{Hap.}} &
        \multicolumn{1}{c}{\textbf{Sad}} &
        \multicolumn{1}{c}{\textbf{Ang.}}
        \rule{0pt}{2ex} \\ \midrule

        \multicolumn{1}{l}{\textbf{IEMOCAP}} & 
        \multicolumn{1}{c}{1099} &
        \multicolumn{1}{c}{947} &
        \multicolumn{1}{c}{608} &
        \multicolumn{1}{c}{289} \rule{0pt}{2ex} \\

        \multicolumn{1}{l}{\textbf{MSP-Improv}} & 
        \multicolumn{1}{c}{2072} &
        \multicolumn{1}{c}{1184} &
        \multicolumn{1}{c}{739} &
        \multicolumn{1}{c}{585} \rule{0pt}{2ex} \\
        
        \bottomrule
    \end{tabular}
    \vspace{-2mm}
    \label{tab:dataset}
\end{table}

\begin{table}
    \centering
    \caption{Ablation study: comparisons between PartialFL and FL on the Food101 data set under different data heterogeneity settings.}
    \begin{tabular}{p{1.5cm}p{1.5cm}p{1.5cm}p{1.5cm}}

        \toprule
        \multicolumn{1}{c}{\textbf{Classification}} & 
        \multicolumn{1}{c}{\textbf{Exp.}} & 
        \multicolumn{1}{c}{\multirow{2}{*}{\textbf{Alpha}}} & 
        \multicolumn{1}{c}{\textbf{Top5}} \rule{0pt}{2.0ex} \\
        
        \multicolumn{1}{c}{\textbf{Modality}} & 
        \multicolumn{1}{c}{\textbf{Setting}} & & 
        \multicolumn{1}{c}{\textbf{Accuracy}} \rule{0pt}{2.0ex} \\

        \midrule

        \multicolumn{1}{c}{\multirow{6}{*}{\textbf{Image}}} & 
        \multicolumn{1}{l}{FL} &
        \multicolumn{1}{c}{\multirow{2}{*}{1.0}} & 
        \multicolumn{1}{c}{51.08\%} \rule{0pt}{2.0ex} \\
        
         & \multicolumn{1}{l}{PartialFL} & &
        \multicolumn{1}{c}{\textbf{53.03\%}} \rule{0pt}{2.0ex} \\
        
        \cmidrule(lr){2-4}
         & \multicolumn{1}{l}{FL} &
        \multicolumn{1}{c}{\multirow{2}{*}{0.5}} & 
        \multicolumn{1}{c}{48.27\%} \rule{0pt}{2.0ex} \\
        
         & \multicolumn{1}{l}{PartialFL} & &
        \multicolumn{1}{c}{\textbf{51.20\%}} \rule{0pt}{2.0ex} \\
        
        \cmidrule(lr){2-4}
         & \multicolumn{1}{l}{FL} &
        \multicolumn{1}{c}{\multirow{2}{*}{0.1}} & 
        \multicolumn{1}{c}{36.20\%} \rule{0pt}{2ex} \\
        
         & \multicolumn{1}{l}{PartialFL} & &
        \multicolumn{1}{c}{\textbf{39.54\%}} \rule{0pt}{2.0ex} \\
        
        \midrule

        \multicolumn{1}{c}{\multirow{6}{*}{\textbf{Image+Text}}} & 
        \multicolumn{1}{l}{FL} &
        \multicolumn{1}{c}{\multirow{2}{*}{1.0}} & 
        \multicolumn{1}{c}{56.04\%} \rule{0pt}{2.0ex} \\
        
         & \multicolumn{1}{l}{PartialFL} & &
        \multicolumn{1}{c}{\textbf{58.62\%}} \rule{0pt}{2.0ex} \\
        
        \cmidrule(lr){2-4}
         & \multicolumn{1}{l}{FL} &
        \multicolumn{1}{c}{\multirow{2}{*}{0.5}} & 
        \multicolumn{1}{c}{52.91\%} \rule{0pt}{2.0ex} \\
        
         & \multicolumn{1}{l}{PartialFL} & &
        \multicolumn{1}{c}{\textbf{55.94\%}} \rule{0pt}{2.0ex} \\
        
        \cmidrule(lr){2-4}
         & \multicolumn{1}{l}{FL} &
        \multicolumn{1}{c}{\multirow{2}{*}{0.1}} & 
        \multicolumn{1}{c}{40.68\%} \rule{0pt}{2ex} \\
        
         & \multicolumn{1}{l}{PartialFL} & &
        \multicolumn{1}{c}{\textbf{43.74\%}} \rule{0pt}{2.0ex} \\

        \bottomrule

    \end{tabular}
    
    \label{tab:non_iid}
\end{table}

\subsection{Models and Features}

\subsubsection{Audio}
The model we use for the SER task is similar to \cite{jaiswal2020privacy}. The network has 2 main components: an embedding network and an emotion classifier. The embedding network of the audio data is a set of convolution layers proposed in \cite{gulati2020conformer}, followed by a Gated Recurrent Unit (GRU) layer \cite{chung2014empirical}. We obtain an utterance-level audio representation by applying mean pooling to the output of the GRU layer. We then use a projection layer to compute the audio embedding. Finally, the emotion classifier takes the audio embedding and estimates emotion labels using a set of dense layers. Inputs to the embedding network are 80 dimensional Mel filter bank (MFB) features using 25ms Hamming window with step size of 10ms. 
We further apply z-normalization to these features within each speaker before feeding them to the embedding network.

\begin{table}
    \caption{Ablation study: Impact of missing modalities in some edge devices; in the Food-101 dataset, the Global model uses Image modality only and we report Top 5 Accuracy; In IEMOCAP and MSP-Improv, we use the Audio only Global model and report UAR.}
    \centering
    \begin{tabular}{p{1.5cm}p{1.5cm}p{1.5cm}p{1.5cm}}
        \toprule
        \multicolumn{1}{c}{\textbf{Dataset}} & 
        \multicolumn{1}{c}{$\boldsymbol \alpha$} & 
        \multicolumn{1}{c}{$\boldsymbol q$} & 
        \multicolumn{1}{c}{\textbf{Scores}} \rule{0pt}{2.0ex} \\
        
        \midrule

        \multicolumn{1}{c}{\multirow{8}{*}{\textbf{Food-101}}} & 
        \multicolumn{1}{c}{\multirow{4}{*}{\textbf{1.0}}} &
        \multicolumn{1}{c}{100\%} & 
        \multicolumn{1}{c}{58.62\%} \rule{0pt}{2.0ex} \\
        
        & &
        \multicolumn{1}{c}{50\%} & 
        \multicolumn{1}{c}{58.04\%} \rule{0pt}{2.0ex} \\
        
        & &
        \multicolumn{1}{c}{25\%} & 
        \multicolumn{1}{c}{57.58\%} \rule{0pt}{2.0ex} \\
        
        & &
        \multicolumn{1}{c}{0\%} & 
        \multicolumn{1}{c}{56.04\%} \rule{0pt}{2.0ex} \\
        
        \cmidrule(lr){2-4}
        
        & \multicolumn{1}{c}{\multirow{4}{*}{\textbf{0.5}}} &
        \multicolumn{1}{c}{100\%} & 
        \multicolumn{1}{c}{51.2\%} \rule{0pt}{2.0ex} \\
        
        & &
        \multicolumn{1}{c}{50\%} & 
        \multicolumn{1}{c}{50.94\%} \rule{0pt}{2.0ex} \\
        
        & &
        \multicolumn{1}{c}{25\%} & 
        \multicolumn{1}{c}{50.31\%} \rule{0pt}{2.0ex} \\
        
        & &
        \multicolumn{1}{c}{0\%} & 
        \multicolumn{1}{c}{48.27\%} \rule{0pt}{2.0ex} \\
        
        \cmidrule(lr){2-4}
        
        & \multicolumn{1}{c}{\multirow{4}{*}{\textbf{0.1}}} &
        \multicolumn{1}{c}{100\%} & 
        \multicolumn{1}{c}{39.54\%} \rule{0pt}{2.0ex} \\
        
        & &
        \multicolumn{1}{c}{50\%} & 
        \multicolumn{1}{c}{39.47\%} \rule{0pt}{2.0ex} \\
        
        & &
        \multicolumn{1}{c}{25\%} & 
        \multicolumn{1}{c}{38.65\%} \rule{0pt}{2.0ex} \\
        
        & &
        \multicolumn{1}{c}{0\%} & 
        \multicolumn{1}{c}{36.20\%} \rule{0pt}{2.0ex} \\
        
        \midrule

        \multicolumn{1}{c}{\multirow{3}{*}{\textbf{IEMOCAP}}} & 
        \multicolumn{1}{c}{\multirow{3}{*}{\textbf{-}}} &
        \multicolumn{1}{c}{100\%} & 
        \multicolumn{1}{c}{61.30\%} \rule{0pt}{2.0ex} \\
        
         & & 
        \multicolumn{1}{c}{50\%} & 
        \multicolumn{1}{c}{59.37\%} \rule{0pt}{2.0ex} \\
        
         & &
        \multicolumn{1}{c}{0\%} & 
        \multicolumn{1}{c}{56.93\%} \rule{0pt}{2.0ex} \\
        
        \midrule

        \multicolumn{1}{c}{\multirow{3}{*}{\textbf{MSP-Improv}}} & 
        \multicolumn{1}{c}{\multirow{3}{*}{\textbf{-}}} &
        \multicolumn{1}{c}{100\%} & 
        \multicolumn{1}{c}{45.14\%} \rule{0pt}{2.0ex} \\
        
         & & 
        \multicolumn{1}{c}{50\%} & 
        \multicolumn{1}{c}{44.97\%} \rule{0pt}{2.0ex} \\
        
         & &
        \multicolumn{1}{c}{0\%} & 
        \multicolumn{1}{c}{43.90\%} \rule{0pt}{2.0ex} \\

        \bottomrule

    \end{tabular}
    
    \label{tab:missing_modality}
\end{table}

\subsubsection{Image}
Similar to the SER model, the image model consists of an embedding network and a classifier. The image embedding network is a set of dense layers followed by a projection layer. We use the image embedding to generate the prediction output. We use the MobileNetV2 \cite{sandler2018mobilenetv2} pre-trained model to extract representations of raw images which are passed to the embedding network. We use this model since it is small enough to fit most edge devices.

\begin{figure*}[h]
    \centering
    \includegraphics[width=0.95\linewidth]{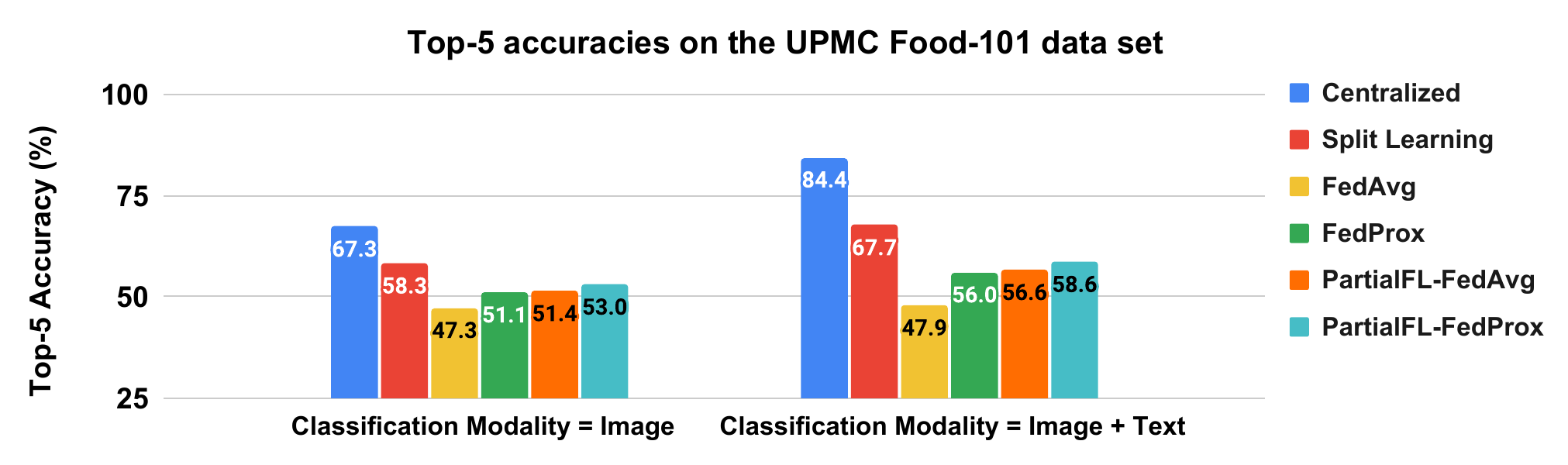}
    \caption{Top-5 accuracies on test set for PartialFL compared to other baselines in the UPMC Food-101 data set. In all cases, we repeat the experiment 5 times with different seed numbers and report average performances. In PartialFL, we report best performance from different temperature values. We use $\alpha = 1.0$ in all federated experiments (FL, SL and PartialFL)}
    \label{img:food101_baseline}
\end{figure*}

\subsubsection{Text}
For the text model, we use an embedding network which maps DistilBert \cite{sanh2019distilbert} sentence representations into a lower-dimensional text embedding. We use a considerably larger embedding network (compared to the edge) in our server. The text model on edge also includes a classifier module (fully connected layer with an outer softmax layer) to generate predictions using local text embeddings. 

\subsubsection{Multi-modal Model}

In the multi-modal global model, we concatenate audio (or image) representations with text representations and pass them through a projection layer to obtain the multi-modal embedding. The classifier then uses this multi-modal embedding to generate the predictions for corresponding tasks.

\section{Additional Results}
\label{sec:additionalresults}

\subsection{Results on UPMC Food-101 dataset}
\label{results:food101}

We report top-1 and top-5 accuracies in this data set. Model performances and sizes for this dataset are shown in Figure \ref{img:food101_baseline}. 

\subsubsection{Uni-modal global model (image only)} Firstly, the centralized image model shows the best performance, largely thanks to a larger model size. Furthermore, we can observe that SL consistently yields better performance than FL. Under FL settings, we find that FedProx outperforms FedAvg, and the top-5 accuracy of FL baseline is about 4\% higher using the FedProx algorithm when compared with FedAvg. PartialFL outperforms the FL baseline by 1.95\% by making use of the shared data modality. SL outperforms PartialFL by only 1.76\%, but at a significant communication overhead. 

\subsubsection{Multi-modal global model (image+text)} Similar to the previous setting, we observe the centralized multi-modal models showing better performance compared to other baselines and FedProx outperforms FedAvg in both FL and PartialFL experiments. Our proposed PartialFL approach improves the global model performance by 2.58\% comparing to the FL baseline.


\subsection{Impact of non-IID settings}

In this experiment, we explore how PartialFL performs under a non-IID setting with data heterogeneity, which is frequently encountered in most real world FL applications. We simulate non-IID edge devices by controlling the $\alpha$ parameter \cite{li2021federated} when creating the data sets, which controls the concentration of the Dirichlet distribution in allocating proportion of label samples to each device (so a small alpha corresponds to more imbalanced data distribution, and hence higher data heterogeneity). We perform this experiment only on the Food-101 data set since the speech data sets are small (with each speaker as a separate edge device). Table \ref{tab:non_iid} presents model comparisons between the PartialFL and FL baselines. We only present results with FedProx, as it showed better performance than FedAvg. As expected, performances of both PartialFL and traditional FL decrease as $\alpha$ value reduces, but PartialFL consistently outperforms FL baselines in all settings. Furthermore, we observe that the PartialFL provides larger performance improvements against FL in non-IID settings (compared to the IID setting of $\alpha=1.0$), by leveraging the shared modality.

\subsection{Impact of missing modalities in edge devices}

In practice, not every edge device may contain multi-modal data; for example, in commercial SLU systems, some devices are designed to only accept audio inputs while others may accept audio, video and text inputs. In this setting, we evaluate performance of PartialFL when the shareable modality is missing in a subset of edge devices. We define $q$ as the percentage of the devices with multi-modal data, and we simulate the case with various values of $q$. $q=0\%$ indicates the case where none of the edge devices contain the shareable modality, and is equivalent to the FL baseline. We explore $q\in\{0\%, 25\%, 50\%, 100\%\}$ and $q\in\{0\%, 50\%, 100\%\}$ in the Food-101 data set and emotion recognition data sets, respectively. 
Results are presented in Table \ref{tab:missing_modality}. In general, we observe that fewer the number of edge devices with the shareable data, lower the performance of the global model. However, this decrease is not substantially large; for example, when $q=25\%$, we find that the top-5 accuracy drops by around 1\% in the Food-101 data set. These results suggest that PartialFL is robust to missing modalities in some devices and the performance decrease is not substantial.

\subsection{Impact of temperature $\tau$}

The temperature parameter $\tau$ in the contrastive loss objectives defines the strength of penalties on the hard negative samples and often has a substantial impact on the final model performance \cite{wang2021understanding}. In this study, we examine its impact on PartialFL. Table \ref{tab:temperature_ablation} shows performances of the uni-modal global model with different temperature parameters in all the datasets we explored. 
As we can see from the table, performance differences at different temperature values are close to each other ($<1\%$) among all values of $\tau$, suggesting that the PartialFL is fairly robust to this hyper-parameter. 

\begin{table}
    \caption{Ablation study: Impact of temperature ($\boldsymbol \tau$) on the PartialFL algorithm.}
    \label{tab:temperature_ablation}
    \centering
    \begin{tabular}{c c c c c}
        \toprule
        \textbf{Model (metric)} & \textbf{Dataset} & \textbf{$\boldsymbol \alpha$ } & \textbf{$\boldsymbol \tau$ } & \textbf{Score} \\ \hline
        \multirow{9}{*}{Image (top-5 acc)} & \multirow{9}{*}{Food-101} & \multirow{3}{*}{1.0} & 0.05 & 53.03\% \\ \cline{4-5} 
         &  & & 0.1 &  52.67\% \\ \cline{4-5} 
         &  & & 0.2 &  51.95\% \\ \cline{4-5}  
         & & \multirow{3}{*}{0.5} & 0.05  & 51.2\% \\ \cline{4-5} 
         &  & & 0.1 &  50.86\% \\ \cline{4-5} 
         &  & & 0.2 &  49.93\% \\ \cline{4-5} 
         & & \multirow{3}{*}{0.1} & 0.05 &  39.42\% \\ \cline{4-5} 
         &  & & 0.1  & 39.54\% \\ \cline{4-5} 
         &  & & 0.2  & 39.38\% \\ \hline
        \multirow{6}{*}{Audio (UAR)} & \multirow{3}{*}{IEMOCAP} & \multirow{3}{*}{-} & 0.05 & 60.71\% \\ \cline{4-5} 
         &  & & 0.1 & 61.30\% \\ \cline{4-5} 
         &  & & 0.2 & 60.52\% \\ \cline{2-5} 
         & \multirow{3}{*}{MSP-Improv} & \multirow{3}{*}{-} & 0.05 & 44.68\% \\  \cline{4-5} 
         & &  & 0.1 & 44.81\% \\  \cline{4-5} 
         &  & & 0.2 & 45.14\% \\ \toprule
    \end{tabular}
\end{table}

\end{document}